\documentclass[letterpaper, 10 pt, conference]{ieeeconf}  

\IEEEoverridecommandlockouts                              
\usepackage{graphics} 
\usepackage{epsfig} 
\usepackage{mathptmx} 
\usepackage{times} 
\usepackage{amsmath} 
\usepackage{amssymb}  
\usepackage{subcaption}  
\usepackage{algorithm}
\usepackage{algpseudocode}
\usepackage{todonotes}
\usepackage{booktabs}
\usetikzlibrary{shapes, arrows}
\usepackage{subcaption}
\usepackage{hyperref}
\usepackage{svg}
\usepackage{algorithm,algpseudocode}

\def\BibTeX{{\rm B\kern-.05em{\sc i\kern-.025em b}\kern-.08em
    T\kern-.1667em\lower.7ex\hbox{E}\kern-.125emX}}

\title{\LARGE \bf
The Persistent Robot Charging Problem for Long-Duration Autonomy
}

\author{Nitesh Kumar$^{1}$, Jaekyung Jackie Lee$^{1}$, Sivakumar Rathinam$^{2}$, Swaroop Darbha$^{2}$, P.B. Sujit$^{3}$ and Rajiv Raman$^{4}$ 
\thanks{Nitesh Kumar was supported in part by the Triad National Security LLC under the grant from the DOE NNSA (award no. 89233218CNA000001), titled ``Modeling and Discrete Optimization Algorithms for Robust Com-
plex Networks''.}
\thanks{$^{1}$ Doctoral student, J. Mike Walker' 66 Department of Mechanical Engineering, Texas A\&M University, College Station, TX - 77843-3123.
        {\tt\small \{niteshk, jkleecontrols\}@tamu.edu}}%
\thanks{$^{2}$ Professor, J. Mike Walker '66 Department of Mechanical Engineering, Texas A\&M University, College Station, TX
        {\tt\small \{srathinam, dswaroop\}@tamu.edu}}%
}

\begin{document}

\maketitle
\thispagestyle{empty}
\pagestyle{empty}

\begin{abstract}

This paper introduces a novel formulation aimed at determining the optimal schedule for recharging a fleet of $n$ heterogeneous robots, with the primary objective of minimizing resource utilization. This study provides a foundational framework applicable to Multi-Robot Mission Planning, particularly in scenarios demanding Long-Duration Autonomy (LDA) or other contexts that necessitate periodic recharging of multiple robots. A novel Integer Linear Programming (ILP) model is proposed to calculate the optimal initial conditions (partial charge) for individual robots, leading to the minimal utilization of charging stations. This formulation was further generalized to maximize the servicing time for robots given adequate charging stations. 
The efficacy of the proposed formulation is evaluated through a comparative analysis, measuring its performance against the thrift price scheduling algorithm documented in the existing literature. The findings not only validate the effectiveness of the proposed approach but also underscore its potential as a valuable tool in optimizing resource allocation for a range of robotic and engineering applications. 

\end{abstract}

\section{INTRODUCTION}

The UAV industry is experiencing rapid expansion, and this technological advancement is immediately felt in everyday life.
An increasing number of companies, such as Amazon, Google, UPS, FedEx, and DHL, are actively integrating UAVs into their operations for last-mile delivery \cite{boysen2018drone}, \cite{liu2021scheduling}. Despite their wide usage, UAVs face limitations in flight time due to their limited battery capacity. For example, a heavy-duty drone like the Prism Lite, which uses 16,000mAh batteries, can achieve only 40 minutes of flight time without any external payload \cite{Prism}.
In Long Duration Autonomy (LDA) and persistent monitoring applications, UAVs are expected to operate over long durations (typically, a few days) without human intervention and hence, this necessitates UAVs to frequently recharge their batteries at a charging station  while carrying out their tasks. 

We assume, without any loss of generality, that no more than one UAV may be recharged at a charging station at any time. Hence, scheduling UAVs for recharging at the charging station becomes pivotal from an operational efficiency viewpoint. Improper scheduling could lead to multiple UAVs running out of charge at the same time. This can happen if any two UAVs have identical charging and flying times and start with the same charge; clearly, the determination of the initial battery charge becomes crucial for staggering/scheduling their deployment, especially in LDA applications requiring persistent monitoring and data collection. Since charging stations are valuable resources that come with significant costs, reducing their number can lead to substantial savings in setup expenses. Therefore, scheduling the charging and departures of UAVs at the charging stations will aid in efficiently utilizing the charging stations and reducing the overall cost.

In this paper, we divide time into uniform time slots (or simply slots) at the charging station; scheduling the charging station implies (a) assigning each time slot to at most one UAV while allowing a UAV to charge fully and (b) making the charging station available for recharging before it runs out of battery charge. We also specify the charging time (time to fully charge starting from no charge state) and the flying time as integer multiples of the time slots; moreover, we call the cycle time for a UAV to be the sum of its charging and flying times.

In this paper, we explore two problem variants: (a) scheduling the UAVs to minimize the number, $m_{min}$ of charging stations, and (b) selecting and scheduling a subset of UAVs to maximize the total flying time of UAVs with given $m \le m_{min}$ charging stations. 

The first variant can be formulated as a generalized non-preemptive windows scheduling problem as discussed by \cite{bar2005windows}, where the objective is to non-preemptively schedule all jobs (or charging UAVs fully before flying and arriving at the charging station before running out of charge) on the fewest possible number of parallel machines (charging stations). This problem has been proven to be NP-hard \cite{bar2005windows}. 

One can relax the generalized non-preemptive scheduling problem by requiring the UAV to arrive at the charging station only when it has completely run out of charge upon reaching the station. This is similar to the thrift price non-preemptive scheduling problem described by \cite{bar2005windows}. Notably, the algorithm proposed by \cite{bar2005windows} offers an 8-approximation solution to the problem. In this paper, we relax the first variant analogous to the thrift price scheduling problem and compare the results with the the 8-approximation approach. Then, we utilize the structure of the solution approach from the first variant to solve the second variant.

Some other existing work in literature such as \cite{shin2019auction,hassija2020scheduling,torky2022scheduling} uses a game-theoretic approach to model the energy trading between UAVs and charging stations in a cost-optimal manner. Their proposed model allocates a time slot to the UAVs using an auctioning process. The above algorithms are designed to work in open groups where there can be any number of participants without any prior knowledge of UAV's parameters such as its charging time and flying time. These methods do not guarantee that each UAV that needs a charging station will have a readily available charging station.

Other recent work on UAVs for LDA applications focuses on optimizing routes while taking into account the energy limitations of UAVs, assuming a sufficient number of available stationary or mobile charging stations. Reference \cite{yu2019algorithms}, \cite{sai_optimal_route}, \cite{sai_bounds}  presents an algorithm to determine the sequence in which different sites and charging stations can be visited. Their algorithm plans tours for unmanned ground vehicles (UGVs) acting as mobile charging stations as well as determining the optimal locations to place stationary charging stations. References \cite{ahmed2016energy,kim2013scheduling} also discuss planning the paths for UAVs using multiple stationary recharging stations. Reference \cite{chour2023agent}  developed an agent-based modeling framework for the multi-UAV rendezvous recharging problem, which consists of energy-limited aerial vehicles that rendezvous with a mobile or fixed charging station. Another recent work, outlined in \cite{2023arXiv230308935B}, addresses a multi-robot persistent monitoring problem involving battery-limited robots. The goal is to determine the minimum number of robots needed to meet latency constraints while also ensuring that the robots periodically recharge at a designated depot. Developing an optimal charging schedule that minimizes the required charging stations can be regarded as a natural follow-on problem of the aforementioned studies. 

The primary novelty and contribution of this paper lies in the development of scheduling strategies for the UAVs at the charging depot. Our first contribution is the introduction of a novel Integer Linear Programming (ILP)-based scheduling algorithm that leverages the cyclic charging and discharging behavior of UAVs and determines their optimal initial charge levels and time slots for charging. These levels are then used to establish a charging schedule, which minimizes the required charging resources at the depot.
Second, in scenarios with limited charging stations, we propose a method for selecting and scheduling UAVs to maximize overall operational efficiency, specifically focusing on maximizing their total total flying time.

Since the application of our proposed method is not limited to UAVs alone, we will use the more general term ``robots" for the remainder of the discussion.

\section{PROBLEM STATEMENT}\label{Prob}

We address the challenge of scheduling charging slots for a fleet of $n$ heterogeneous robots at a charging depot equipped with $m$ number of distinct charging stations. The primary constraint is to ensure that each robot requiring charge obtains a charging station while avoiding running out of battery charge. Specifically, a robot must have an empty charging station ready after completing its mission; otherwise, it is not deployed. The operational parameters for each robot include the charging time and the operational time. It is assumed that the specifications for each robot i.e., the operational time and the necessary charging time to recharge from empty to full are known beforehand.

One of the main requirements for efficient scheduling is to guarantee the availability of a charging station for each robot in need over an infinite time horizon. However, if we can pose our problem as a periodic problem then the scheduling problem is periodic with a cycle time, say $T$, then we need to satisfy the aforementioned constraint only until time $T$.  Each robot is allocated consecutive charging slots equivalent in duration to its specified charging time. After charging fully, we assume the robot will require the charging station again exactly after its total operational time, establishing a periodic problem framework. Notably, partial charging and partial operation are not permissible, except during the initial deployment.

A feasible solution involves using at least as many charging stations as the number of robots, i.e. $m\geq{n}$; then, it guarantees the availability of a charging station for each robot when needed. However, that might not be an efficient usage of the charging stations because  the corresponding charging station will be left unutilized whenever a robot is in operation. This setup raises two critical questions:

\begin{enumerate}
\item Does there exist a charging schedule that maximizes the resource utilization by minimizing the required resources \(m\) for a given set of \(n\) heterogeneous robots?
\item How can we construct an optimal schedule that maximizes the total flight time of the chosen set of  robots from a fleet of \(n\) when limited by \(m\) resources?
\end{enumerate}

In the next section, we will address the two questions raised above using this non-preemptive approach.

\section{SCHEDULING FRAMEWORK}\label{sol}
In subsection \ref{time_period}, we discuss the periodic nature of scheduling and the associated scheduling horizon (period); we also propose a method to determine the cycle time that can serve as the scheduling horizon for optimizing the charging schedule. In subsection \ref{Min_m}, we delve into the methodology for determining the optimal charging schedule for each robot that minimizes the required charging stations. In subsection \ref{Max_f}, we outline a strategy for selecting robots from a given set and scheduling them optimally on limited charging stations to maximize their cumulative operational time across the scheduling horizon. 

\subsection{Finding a Scheduling Horizon}\label{time_period}

Let $C$ and $D$ respectively denote the sets of charging and flying times for the robots, i.e., 
\begin{equation} \label{charging}
C=\{c_1,c_2,\hdots,c_n\}, \;\; c_i \in \mathbb{Z^+}, \;\; i=1,2,\dots,n,
\end{equation}
\begin{equation} \label{discharging}
D=\{f_1,f_2,\hdots,f_n\}, \;\; f_i \in \mathbb{Z^+}, \;\; i=1,2,\dots,n.
\end{equation}
The charging time $c_i$, and operational time $f_i$ are assumed to be integers and are known a priori.

For each robot $i$, we consider a sequence of actions consisting of charging for $c_i$ slots followed by discharging for $f_i$ slots resulting in a cycle time $T_i=c_i + f_i$ slots. This cycle time $T_i$ represents the interval after which the robot returns to its initial state. Therefore, the process of charging and discharging is periodic for each robot with a cycle time $T_i$. 

If each robot begins with an initial charge and operates on an individual cycle time $T_i$, and assuming there are enough charging stations available, all robots will return to their initial conditions simultaneously after $T$ slots, where $T$ represents the least common multiple (LCM) of the individual cycle times of the robots,

\begin{equation} \label{Time period}
 T=\operatorname{LCM}(T_1,T_2,\hdots,T_n).
\end{equation}

The non-preemptive approach simplifies the problem to a periodic problem, whose cycle time is given by Eq. (\ref{Time period}), irrespective of the initial conditions of the robots. This information can be utilized to reduce the optimization problem from an infinite time horizon problem to the time with scheduling horizon $T$. 

\subsection{Minimizing resources}\label{Min_m}

In this subsection, we present a novel formulation that will aid the determination of the initial conditions (charge) of the robots and find the corresponding minimum number of charging stations required over the scheduling horizon $T$. Here, the initial condition represents the amount of charge available with the robot and whether it is charging or in operation at the moment. 

\begin{figure}[htb!]
  \centering
  \begin{subfigure}{0.45\linewidth}
    \centering
    \includegraphics[width=\linewidth]{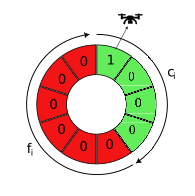}
    \caption{State at time $t$, illustrating the initiation of charging for this specific example.}
    \label{fig:sub1}
  \end{subfigure}%
  \hspace{0.5em} 
  \begin{subfigure}{0.45\linewidth}
    \centering
    \includegraphics[width=\linewidth]{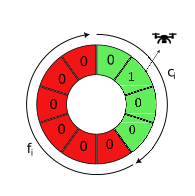}
    \caption{State at time $t+1$, showcasing the robot's completion of 1 slot of charging.}
    \label{fig:sub2}
  \end{subfigure}
  \caption{Evolution of robot's states: Illustration of the charging (green) and discharging (red) phases as the robot transitions from one state to the next.}
  \label{fig:1}
\end{figure}

 One can associate a time wheel as shown in Fig.\ref{fig:1} that helps visualize the periodic evolution of charging and discharging of the robot's battery. Fig.\ref{fig:1} depicts four charging slots and six operational slots with a cycle time of ten slots for a robot. The state evolves in a clockwise direction along the time wheel as time increases. For example, from Fig.1(a), we know that the robot has just begun charging at time $t$, and from Fig.1(b), it is in the second slot of charging at time $t+1$. The depiction of the number $1$ in the time wheel indicates the state of the robot's battery and its status (charging/discharging). 

\begin{figure}[htb!]
\vspace{0.25cm}
  \centering
    \includegraphics[width=\linewidth]{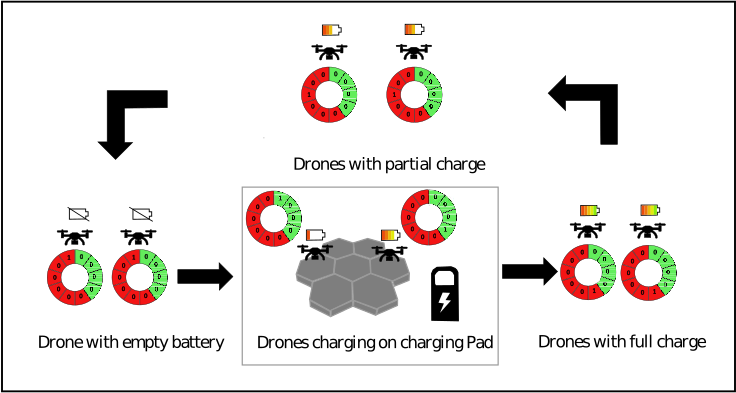}
    \caption{State Transition Schematic: Depiction of the cyclic evolution of robot's states throughout the charging cycle. The transition initiates with charging, indicated by the green segment, progressing clockwise. Upon reaching full charge, the state transitions clockwise into the red segment.}
    \label{fig:4}
\end{figure}

Reflecting the state of the battery charge as depicted in the time wheel depicted above Fig.\ref{fig:1}, we associate a {\it binary} cyclic state vector \(\mathbf{r_i(t)}\) of size $T_i$ which contains exactly one component that is a ``$1$''. This constraint can be captured through the equation:
\begin{align}
\label{bin_r}
   {\mathbf 1}^{T_i} {\mathbf r}_i(t) = 1, \; t \ge 0, 
\end{align}
where ${\mathbf 1}$ is a {\it row} vector of the size $T_i$ with all its components being $1$. The periodic state transition of ${\mathbf r}_i(t)$ can be described by:
\begin{equation}\label{r_it}
\mathbf{r_i(t)}= \mathbf{A_i} \mathbf{r_i(t-1)}, \quad t \ge 1
\end{equation}
where
\begin{equation} \label{state transition matrix}
     \mathbf{A_i} =
    \begin{bmatrix}
        0 & 0 & \hdots & 0 & 1 \\
        1 & 0 & \hdots & 0 & 0 \\
        0 & 1 & \hdots & 0 & 0 \\
        \vdots & \vdots & \hdots & \vdots & \vdots \\
        0 & 0 & \hdots & 1 & 0
    \end{bmatrix}.
\end{equation}

The state transition matrix $A_i$  is a permutation matrix and is also circulant. The state vector $r_i(t)$ at time $t$ can be used to determine whether the robot is recharging (and if so, it will a charging station is required for this purpose); this is done through a charging indicator vector $\mathbf{p^\intercal_i}$. The  first $c_i$ elements  of the vector
\(\mathbf{p_i}\) are $1$ and remaining $f_i$ elements are $0$; this vector captures the charging and discharging characteristics of the $i^{th}$ robot's battery. A binary decision variable $z_i(t)$ indicates whether the robot $i$ is utilizing the resource (charging station) or not at time $t$ through: 
\begin{equation}\label{z_it}
z_i(t)=\mathbf{p^\intercal_i}\mathbf{r_i(t)}.
\end{equation}

From this discussion, we can now formulate an Integer Linear Program (ILP) to find the initial condition \(\mathbf{r_i(0)}\) to determine the  minimum number, $m_{min}$, of charging stations for scheduling over scheduling horizon $T$.  Mathematically,



\begin{equation} \label{charging station}
m_{min}=\min\;\;{m},
\end{equation}
subject to the following constraints \eqref{bin_r}, \eqref{r_it} and 
\begin{equation} \label{state transition}
    \begin{bmatrix}
        \sum_{i=1}^{n} {z_i(1)}\\
       \sum_{i=1}^{n} {z_i(2)} \\
        \vdots \\
       \sum_{i=1}^{n} {z_i(T)}\\
    \end{bmatrix}
     = \begin{bmatrix}
       \mathbf{p^\intercal_1} &  \hdots & \mathbf{p^\intercal_n} \\
        \mathbf{p^\intercal_1A_1} &  \hdots & \mathbf{p^\intercal_nA_n} \\
        \vdots & \hdots & \vdots \\
        \mathbf{p^\intercal_1}\mathbf{A_1}^{T-1} & \hdots & \mathbf{p^\intercal_n}\mathbf{A_n}^{T-1}\\
    \end{bmatrix}
    \begin{bmatrix}
        \mathbf{r_1(0)}   \\
        \mathbf{r_2(0)}    \\
        \vdots\\
        \mathbf{r_n(0)}   \\
    \end{bmatrix}
    \leq \begin{bmatrix}
        m\\
        m \\
        \vdots\\
        m\\
    \end{bmatrix}.
\end{equation}
The last set of constraints indicates that no more than $m$ resources (charging stations) be utilized at any time.

The above-formulated problem ILP can be solved for the initial conditions \(\mathbf{r_i(0)}\), and the minimum required resources $m_{min}$.  Leveraging the obtained initial conditions \(\mathbf{r_i(0)}\) and the relationship expressed in 
\begin{equation}\label{z}
z_i(t)=\mathbf{p^\intercal_i} \mathbf{A_i}^t \mathbf{r_i(0)}.   
\end{equation}
one can determine the charging schedule for each robot.

\subsection{Maximizing flying time}\label{Max_f}

In this section, we establish an optimal schedule for deploying robots by selecting which robots to operate and initializing their battery states. Unlike in the previous section, we assume specific knowledge of the number of available resources, denoted as \(m \leq m_{min}\). The objective of this problem is to maximize the total operational time of all robots within the final schedule, given a finite number of resources \(m\).

A robot, once chosen, is utilized throughout the entire mission; if not selected, it is not deployed. This decision is represented by a binary variable \(u_i\), where \(u_i = 1\) if the \(i^{th}\) robot is deployed, and \(u_i = 0\) otherwise.

Let us define a vector \(\mathbf{q}_i = [ {\mathbf 0}^{c_i}\  {\mathbf 1}^{f_i} ]^T \) that can be used to determine if the $i^{th}$ robot is in operation or charging. The vector $\mathbf{q}_i$ is specific to the $i^{th}$ robot as $c_i, f_i$ represent its charging and operation times respectively.
For instance, if the \(i^{th}\) robot has a charging time \(c_i = 3\) and a operation time \(f_i = 5\), the corresponding vector is \(\mathbf{q}_i^T = [ {\mathbf 0}^{3}\ {\bf 1}^{5} ] = [ 0\ 0\ 0\ 1\ 1\ 1\ 1\ 1 ]\).  Define a binary variable $y_i(t)$ that indicates whether the robot is in operation or not, depending on whether it is $1$ or $0$:
\begin{equation} 
y_i(t) = \mathbf{q}_i^T  \mathbf{r_i(t)} = \mathbf{q}_i^T\mathbf{A}_i^t  \mathbf{r_i(0)}.
\end{equation}


In this scenario, our objective is to maximize the total operation time of all robots and across the time horizon:

\begin{equation} 
\max \sum_{i=0}^{T-1} \sum_{i=1}^{n} y_i(t),
\end{equation}
subject to the following constraints:

\begin{equation} 
 \begin{bmatrix}
       \mathbf{p^\intercal_1} &  \hdots & \mathbf{p^\intercal_n} \\
        \mathbf{p^\intercal_1A_1} &  \hdots & \mathbf{p^\intercal_nA_n} \\
        \vdots & \hdots & \vdots \\
        \mathbf{p^\intercal_1}\mathbf{A_1}^{T-1} & \hdots & \mathbf{p^\intercal_n}\mathbf{A_n}^{T-1}\\
    \end{bmatrix}
    \begin{bmatrix}
        \mathbf{r_1(0)}   \\
        \mathbf{r_2(0)}    \\
        \vdots\\
        \mathbf{r_n(0)}   \\
    \end{bmatrix}
    \leq \begin{bmatrix}
        m\\
        m \\
        \vdots\\
        m\\
    \end{bmatrix},
\end{equation}

\begin{equation}
\begin{bmatrix}
{\mathbf 1}^{T_1} & {\mathbf 0}^{T_2} & \cdots & {\mathbf 0}^{T_n} \\
{\mathbf 0}^{T_1} & {\mathbf 1}^{T_2} & \cdots & {\mathbf 0}^{T_n} \\
\vdots & \vdots & \ddots & \vdots \\
{\mathbf 0}^{T_1} & {\mathbf 0}^{T_2} & \cdots & {\mathbf 1}^{T_n}
\end{bmatrix}
    \begin{bmatrix}
        \mathbf{r_1(0)}   \\
        \mathbf{r_2(0)}    \\
        \vdots\\
        \mathbf{r_n(0)}   \\
    \end{bmatrix}
= \mathbf{W} = 
\begin{bmatrix}
u_1 \\
u_2 \\
\vdots \\
u_n
\end{bmatrix}.
\end{equation}

The vectors \( \mathbf{r}_i(0) \) and \( \mathbf{W} \) are to be determined for each robot to optimize the desired operational outcome, which in this case, is the maximization of the operational time for the entire fleet.

\begin{figure}[htb!]
  \centering
    \includegraphics[width=\linewidth]{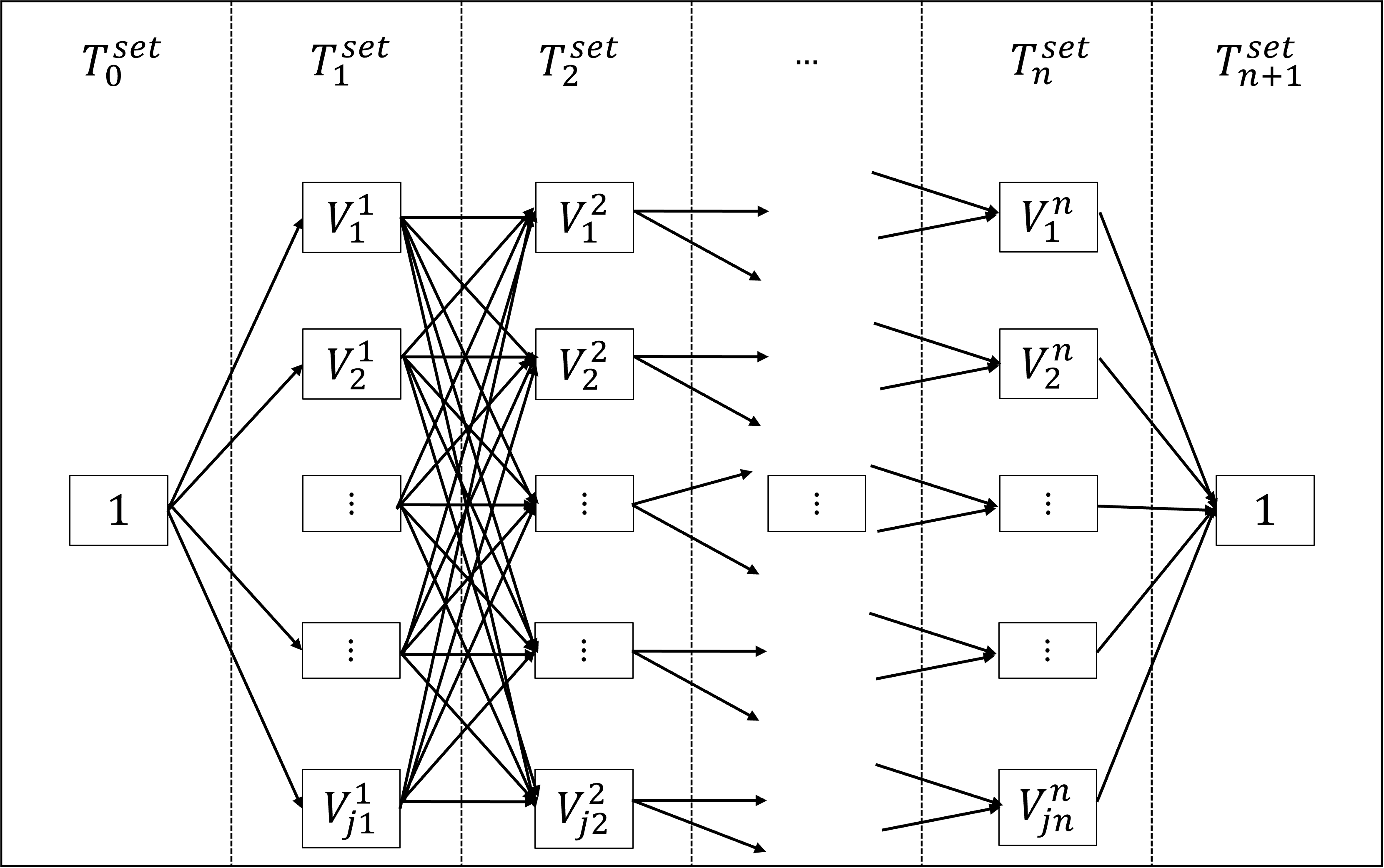}
    \caption{Graph representing the candidate cycle times}
    \label{Dijsktra}
\end{figure}

\section{ROBUST SCHEDULING}\label{rob}

From a practical perspective, maintaining a reserve of fuel is essential for robots to account for unforeseen delays. This reserve ensures that, even if a robot is delayed, it can still reach the charging station. In Subsection \ref{Safety Margin}, we explore the concept of safety margins, which function not only as a fuel reserve but also have the potential to reduce the scheduling horizon. This reduction in the scheduling horizon can, in turn, decrease computational complexity and simplify the overall scheduling process. Subsequently, in Subsection \ref{managing delay}, we discuss a strategy for managing delayed robots without disrupting the schedules of other robots.

\subsection{Safety margins}\label{Safety Margin}
The cycle time $T_i$ of the $i^{th}$ robot is defined as $T_i=f_i+c_i$ and, as discussed in Section \ref{time_period}, the overall cycle time (scheduling horizon) $T$ is calculated as $T=\operatorname{LCM}(T_1,T_2,\hdots,T_n)$. However, $T$ can potentially be large if  $T_i$'s are co-prime. Having a large $T$ implies a longer scheduling horizon, leading to increased computational complexity of determining the charging schedule; this may not be desirable in certain applications. A shorter scheduling horizon allows for tactical flexibility.

To reduce $T$ and achieve a lower scheduling horizon, we propose a method inspired by Dijsktra's shortest path algorithm. This approach involves leveraging the fact that $T$ is the LCM of the individual robots's cycle times. By modifying the cycle time $T_i$ of individual robots, $T$ can be controlled. This modification can be achieved in two ways:  a) by allowing the robot to wait at the charging station, which increases its cycle time, or b) by limiting the robot's operational time, which decreases its cycle time and provides a safety margin for the robot.

Increasing the cycle time through waiting can help reduce the scheduling horizon but also increases the number of variables, potentially increasing the computational time for finding an optimal schedule. Conversely, reducing the operational time decreases the cycle time, the number of variables and the scheduling horizon; ultimately, it results in a lower computation time. We adopt the latter approach here.


We relax the operational time for the $i^{th}$ robot to belong to a set, with each candidate operational time in the set bounded by $f_i$. The problem thereafter reduces to finding an element in each set so that the LCM of their corresponding cycle times is the smallest. 

We refer to each robot as a node and the candidate cycle times for that robot as vertices of that node. Using these candidate cycle times as vertices, we create a weighted graph, as illustrated in Fig.\ref{Dijsktra} where each vertex of $i^{th}$ node is connected to each vertex of $(i+1)^{th}$ node. We then apply the idea of Dijkstra's shortest path to find the path that minimizes the LCM of the chosen vertices from each node.


The candidate cycle time, $T^c_i$, for the $i^{th}$ robot is:
\begin{equation}\label{new_time}
    T^c_i = f^{new}_i+c_i \;\;,\;\;f^{new}_i\leq f_i,\;\; f^{new}_i \in \mathbb{Z^+}, \;\;\text{for}\;\; i=1,2,\dots,n.
\end{equation}
Here, $c_i$ (charging time) is fixed and known in advance, while $f^{new}_i$ (operational time) is a variable that can be adjusted. 

A control parameter, $\epsilon$, further restricts the set, $T_i^{set}$, to which $T^{c}_i$ can belong: 
\begin{equation} \label{charging}
T_i^{set}=\{V \; : \;(1-\epsilon_i)T_i\leq V\leq T_i  \;\; \text{and}  \;\; V \in \mathbb{Z^+} \},\;\; \forall\;\; i=1,2,\dots,n.
\end{equation}
The parameter $\epsilon$ represents the maximum fraction of residual battery charge when the robot arrives at the charging station for recharging. 

We then construct a graph (shown in Fig.\ref{Dijsktra}) using these candidate cycle times $V^i_j$, where each vertex $V^i_{k_1}$ is connected to $V^{i+1}_{k_2}$ where $V^i_{k_1}\in T_i^{set} \text{ and } V^{i+1}_{k_2}\in T_{i+1}^{set}$. Additionally, we introduce two pseudo vertices at the start and end of the graph with a value of 1.

The cost associated with choosing a path through vertices $V^1_{k_1},V^2_{k_2},\dots,V^n_{k_n}$ is defined as $LCM(V^1_{k_1},V^2_{k_2},\dots,V^n_{k_n})$ instead of the conventional sum of edge lengths. This approach allows us to reduce the scheduling horizon $T$ by adjusting the operational times $f^{new}_i$ for each robot while controlling the maximum safety margin using $\epsilon_i$.

\subsection{Managing Delay}\label{managing delay}
This section addresses scenarios where a robot does not arrive at the charging station at its scheduled time slot. There are two possible situations: a) The robot arrives at the depot before its scheduled slot. In this case, the robot can wait in the waiting area until its turn comes, or b) the robot arrives later than its scheduled slot. In the second case, accommodating the tardy robot can disrupt the schedule of other robots. Since it arrived late, the tardy robot will release the resource (depart from the charging station) at a later time, causing the next robot in line to wait. This can result in a cascading effect of delays, where each subsequent robot must wait at the charging station for its turn, leading to changes in the schedule of every robot. 

From Eq.(\ref{z}), we can conclude that the charging schedule of a robot is solely a function of its initial conditions and physical parameters. Since this robot has missed its deadline to avail the charging station, we need to find an alternative charging schedule or initial condition that guarantees the earliest possible charging slot without disrupting the schedules of other robots.

Let us assume that the $k^{th}$ robot misses the deadline. From our earlier discussion from section \ref{sol} we can safely assume that the required charging stations $m$ and the corresponding initial condition $\mathbf{r_i(0)} \;\forall\; i \in \{1,2,\hdots,n\}\;\backslash
\{k\}$  (and correspondingly, their schedules) have already been computed. Using this information, we want to find the new initial condition \(\mathbf{r^*_k(0)}\) for the $k^{th}$ robot. 

Eq.\eqref{state transition} can be re-written as follows:

 \begin{align} 
    \begin{bmatrix}
        \mathbf{p_k}^\intercal  \\
        \mathbf{p_kA_k}^\intercal  \\
        \vdots  \\
        \mathbf{p_k}^\intercal\mathbf{A_k}^{T-1}\\
    \end{bmatrix}\mathbf{r^*_k(0)}
    &\leq 
    \begin{bmatrix}
        m\\
        m \\
        \vdots\\
        m\\
    \end{bmatrix}-
    \sum_{\substack{i=1 \\ i \neq k}}^{n}\begin{bmatrix}
        \mathbf{p_i}^\intercal  \\
        \mathbf{p_iA_i}^\intercal  \\
        \vdots  \\
        \mathbf{p_i}^\intercal\mathbf{A_i}^{T-1}\\
    \end{bmatrix}\mathbf{r_i(0)}, \\
    {\mathbf 1}^{T_k} {\mathbf r}_k^*(0) &= 1.
\end{align}

\begin{algorithm}
\caption{Algorithm to reduce the scheduling horizon}
\begin{algorithmic}[1]
\Function{dijkstra\_lcm}{graph, src, target}

    \State \textit{cost} $\gets$ array of size $n+2$ filled with $\infty$
    \State \textit{cost[src]} $\gets 1$
    \State \textit{pq} $\gets$ priority queue initialized with $(1, \text{src})$
    \State \textit{visited} $\gets$ array of size $n+2$ filled with \textit{false}
    \State \textit{predecessors} $\gets$ array of size $n+2$ filled with $-1$

    \While{\textit{pq} is not empty}
        \State Sort \textit{pq} in descending order
        \State $(\text{min\_cost}, u) \gets$ pop last element from \textit{pq}
        
        \If{\textit{visited}[$u$]}
            \State \textbf{continue}
        \EndIf
        
        \If{$u$ = \textit{target}}
            \State \textit{path} $\gets [u]$
            \While{\textit{predecessors}[$u$] $\neq -1$}
                \State Append \textit{predecessors}[$u$] to \textit{path}
                \State $u \gets \textit{predecessors}[u]$
            \EndWhile
            
            \State \textbf{return} \textit{min\_cost}, \textit{reverse(path)}
            
        \EndIf
        
        \State \textit{visited}[$u$] $\gets$ \textit{true}
        
        \For{each $(v, \text{weight})$ in \textit{graph}[$u$]}
            \State \textit{new\_cost} $\gets \text{lcm(min\_cost, weight)}$
            \If{\textit{new\_cost} $<$ \textit{cost}[$v$]}
                \State \textit{cost}[$v$] $\gets \textit{new\_cost}$
                \State Push $(\textit{new\_cost}, v)$ to \textit{pq}
                \State \textit{predecessors}[$v$] $\gets u$
            \EndIf
        \EndFor
    \EndWhile
\EndFunction
\end{algorithmic}
\end{algorithm}

The above set of inequalities is always feasible, as the original schedule is feasible; in this case, the tardy robot will have to wait until its turn for recharging comes again in the original schedule. In order to minimize the wait time, one can find the set of all feasible $r_k^{*}(0)$ that satisfy the above inequalities, and pick the one with the smallest wait time.



\section{COMPUTATIONAL RESULTS}\label{com}

 This section is divided into two parts:  In subsection \ref{comp_min_m}, we compare our proposed resource minimization methodology with existing approaches. In subsection \ref{Scheduling_Horizon}, we present the results of reduced scheduling horizon. The simulations were executed on a computational platform comprising an Ubuntu 20.04 NUC computer, equipped with an Intel i7-6770HQ processor and 32GB of RAM. All simulation data is available at the following  \href{https://github.com/Nitesh-mk/Persistent-Scheduling-Problem-Data-Set.git}{GitHub Repository}.

\subsection{Minimizing  resources}\label{comp_min_m}

We conducted a comparative analysis between our proposed Integer Linear Programming-based Scheduling (ILPS) method and the Thrift Price Window Scheduling (TPWS) method, as described in \cite{bar2005windows}. The TPWS algorithm is considered optimal when the charging time $c_i$ (job processing time) and cycle time $T_i$ (window) are powers of 2. For more general cases, TPWS creates a schedule by rounding the charging time $C_i$ up to the nearest power of 2 and the cycle time $T_i$ down to the nearest power of 2. In such scenarios, TPWS is proven to be an 8-approximation algorithm \cite{bar2005windows}.

To evaluate the effectiveness of our proposed ILPS algorithm, we conducted simulations under two scenarios:  (a) when all $c_i$ and $T_i$ are powers of 2, and (b) where $c_i$ and $T_i$ deviate from powers of 2. In Case (a), both TPWS and ILPS provide exactly same optimal solution, highlighting ILPS’s ability to achieve
optimal solutions. Case (b) is created by adding slight perturbations to Case (a) and ideally the results should not deviate drastically between the 2 scenarios.   ILPS demonstrates effective results similar to case (a), unlike TPWS which fails to deliver optimal results where $c_i$ and $T_i$ are not powers of 2.

For both cases, we analyze how the number of charging stations required to schedule is affected by the number of robots available for deployment and the scheduling horizon $T=\operatorname{LCM}(T_1,T_2,\hdots,T_n)$. 

\begin{figure}[hbt!]
\vspace{0.25cm}
  \centering
  \begin{subfigure}{0.48\linewidth}
    \centering
    \includegraphics[width=\linewidth]{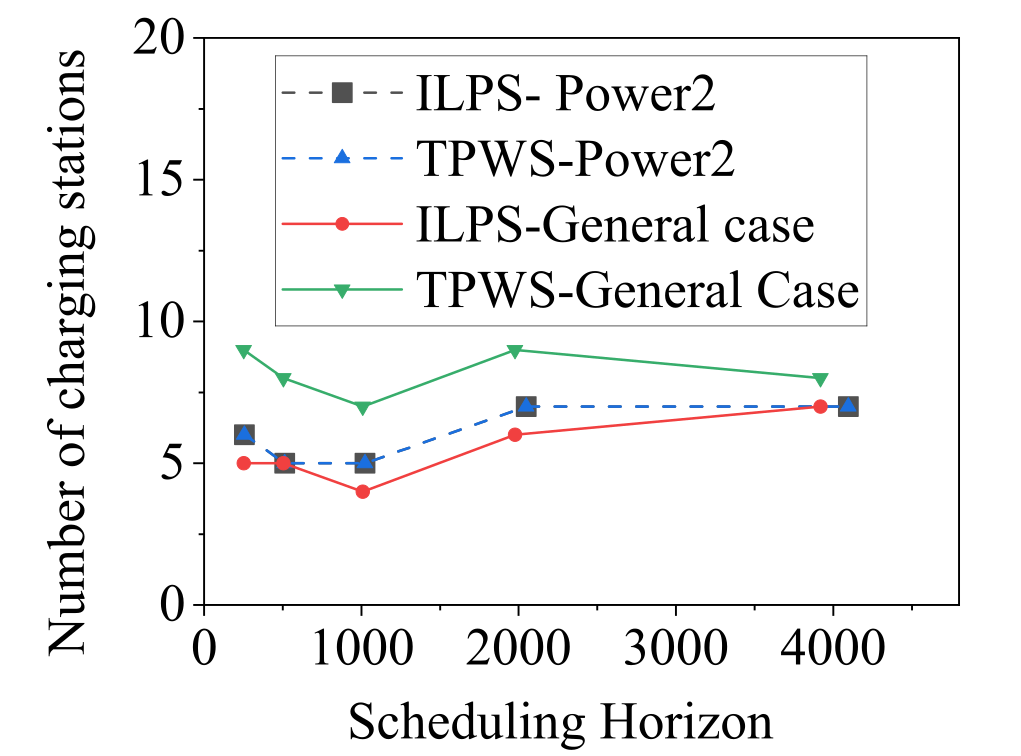}
    \caption{Optimal $\#$ of Charging Stations vs Scheduling Horizon}
    \label{fig:sub_21}
  \end{subfigure}%
  \hfill
  \begin{subfigure}{0.48\linewidth}
    \centering
    \includegraphics[width=\linewidth]{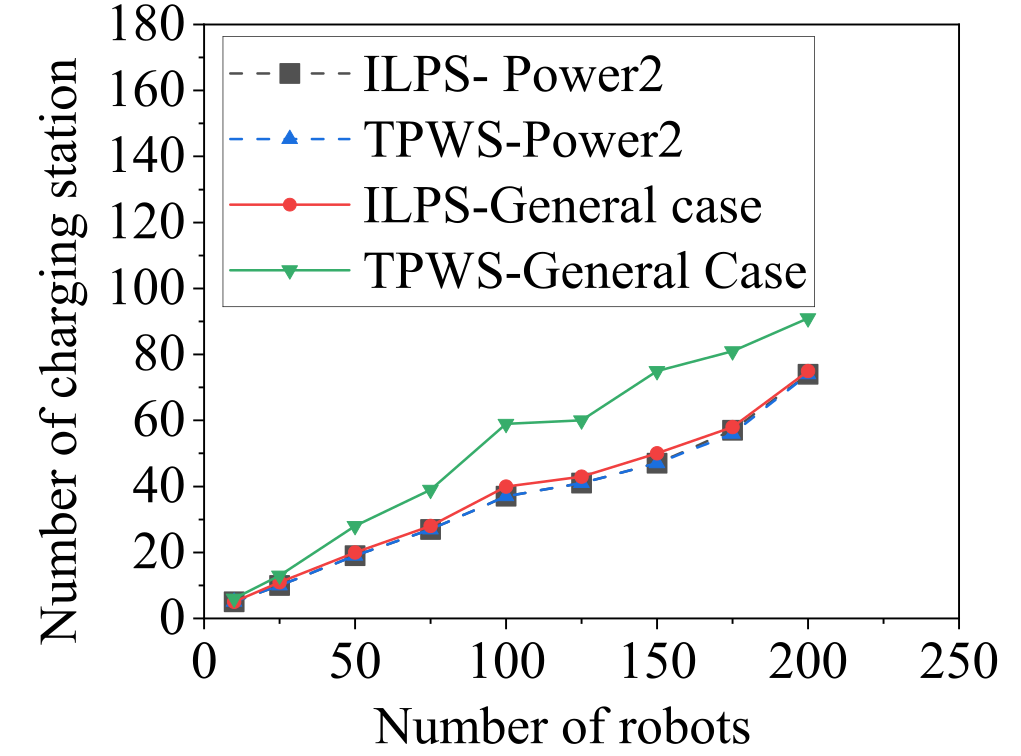}
    
    \caption{Optimal $\#$ of Charging Stations  vs  $\#$ robots}
    \label{fig:sub_22}
  \end{subfigure}
  \caption{Performance of the Integer Linear Programming-based Scheduling (ILPS) algorithm and Thrift Price Window Scheduling (TPWS) algorithm.}
  \label{fig:2}
\end{figure}

The simulation employs a set of 10 heterogeneous robots. Fig.\ref{fig:sub_21} presents the results of the two algorithms as the scheduling horizon increases. Notably, TPWS and ILPS yield optimal results for instances where the scheduling horizon and charging time are powers of 2. However, as we slightly perturb the parameters from these power-of-2 instances, TPWS deviates from the optimal solution, while ILPS consistently identifies the optimal solution. Fig \ref{fig:sub_22} further demonstrates this comparison as the number of robots increases. For power-of-2 instances, ILPS aligns with TPWS, while the deviation grows in TPWS as the number of robots increases. These findings underscore the robustness of the ILPS algorithm, particularly in scenarios where TPWS encounters challenges in maintaining optimality.

\begin{figure}[htb!]
    \centering
    \begin{subfigure}{0.48\linewidth}
        \includegraphics[width=\linewidth]{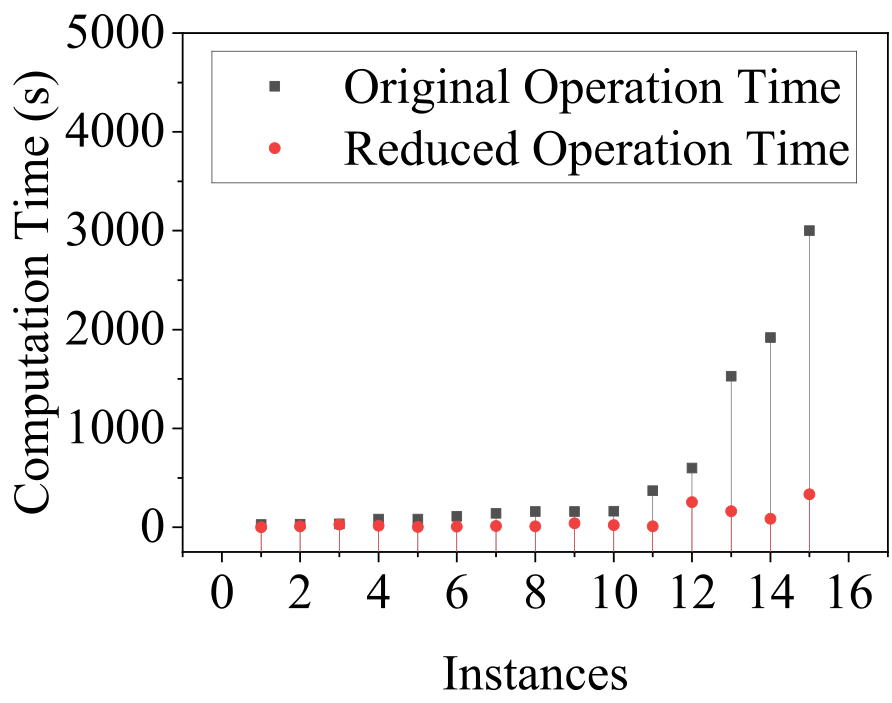}
        \caption{Computation Time}
        \label{Computation Time}
    \end{subfigure}
    \hfill
    \begin{subfigure}{0.48\linewidth}
        \includegraphics[width=\linewidth]{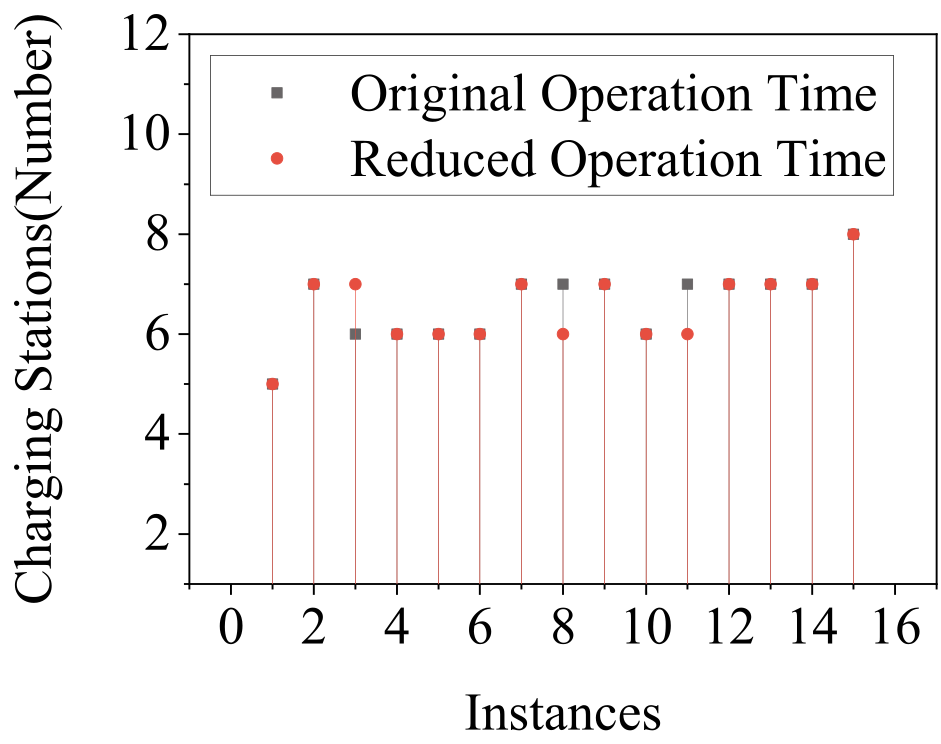}
        \caption{Charging Stations}
        \label{fig:subim2}
    \end{subfigure}
    \caption{Comparison between Original Operational Time and Reduced Operational Time. }
\end{figure}

\begin{figure*}[htb!]
\vspace{0.25cm}
    \centering
        \includegraphics[width=\linewidth]{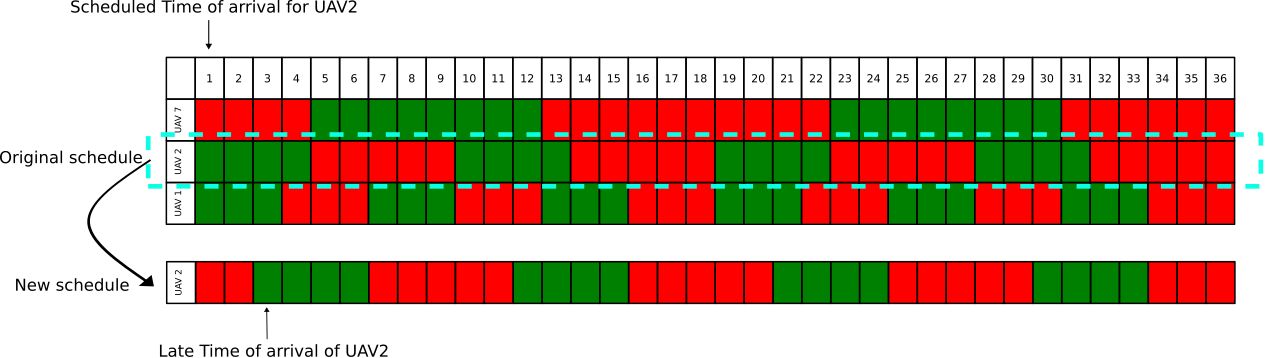}
        
    \caption{Schedule of the robots for the two cases: (a) when all robots are on time, and (b) when UAV2 is delayed. In both schedules, green represents the charging time and red represents the flying time. }\label{fig:charging_schedule}
\end{figure*}

\begin{figure*}[hbt!]
    \centering
    \begin{subfigure}{0.48\textwidth}
        \includegraphics[width=\linewidth]{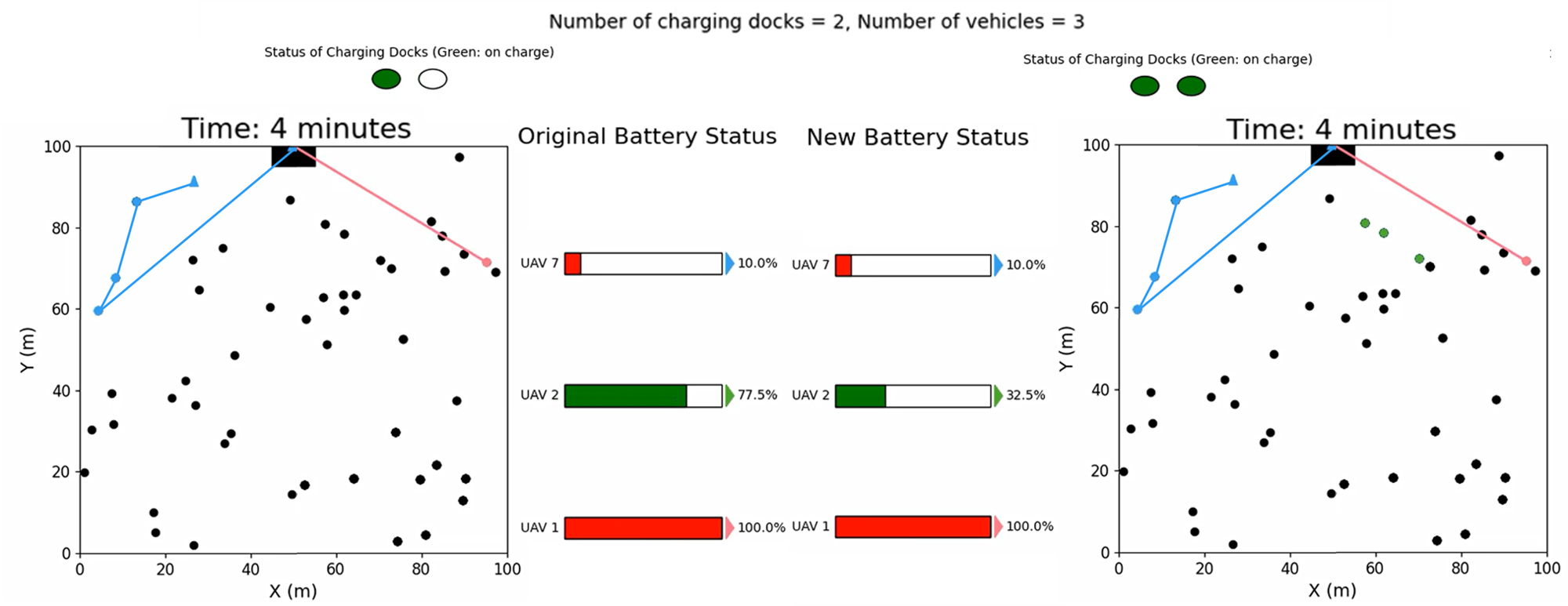}
        \caption{Snapshot at 4 minutes: Robots 1 and 7 continue their missions, while 2 recharging}
        \label{fig:subim1}
    \end{subfigure}
    \hspace{5pt}
    \begin{subfigure}{0.48\textwidth}
        \includegraphics[width=\linewidth]{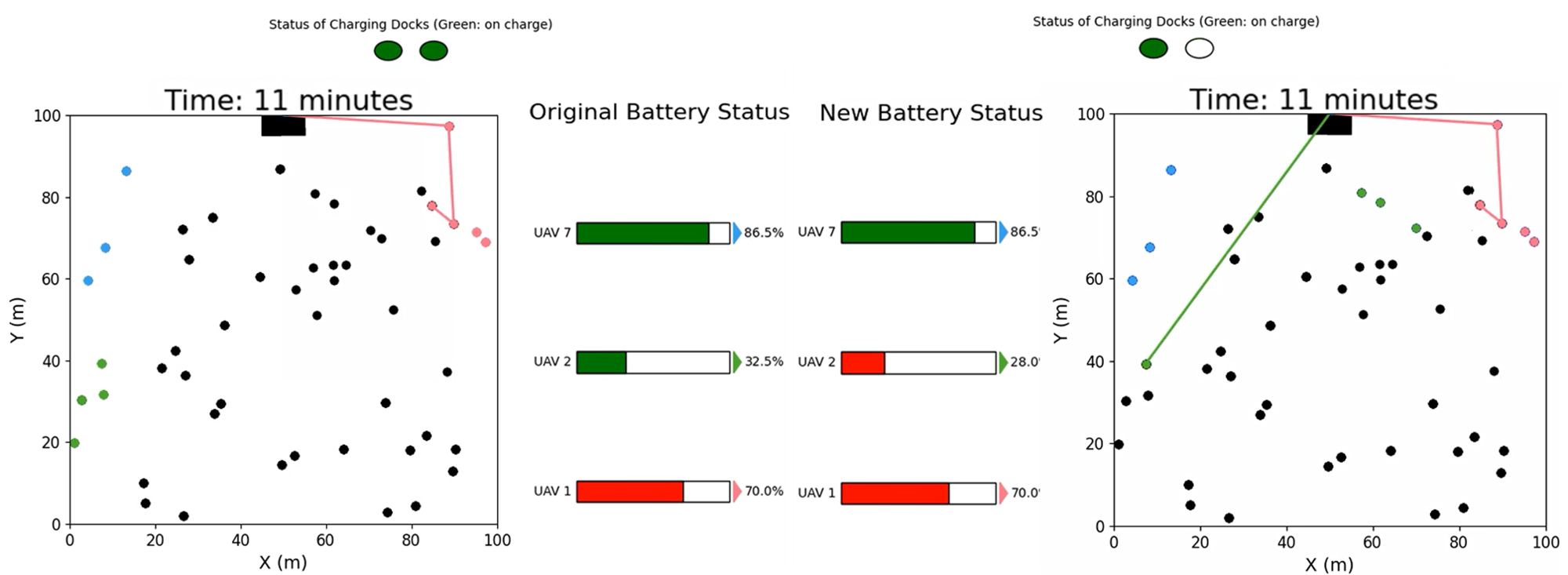}
        \caption{Snapshot at 11 minutes: Robots 1 and 2 continue their missions, while 7 recharging}
        \label{fig:subim2}
    \end{subfigure}
    \hspace{5pt}
    \begin{subfigure}{0.48\textwidth}
        \includegraphics[width=\linewidth]{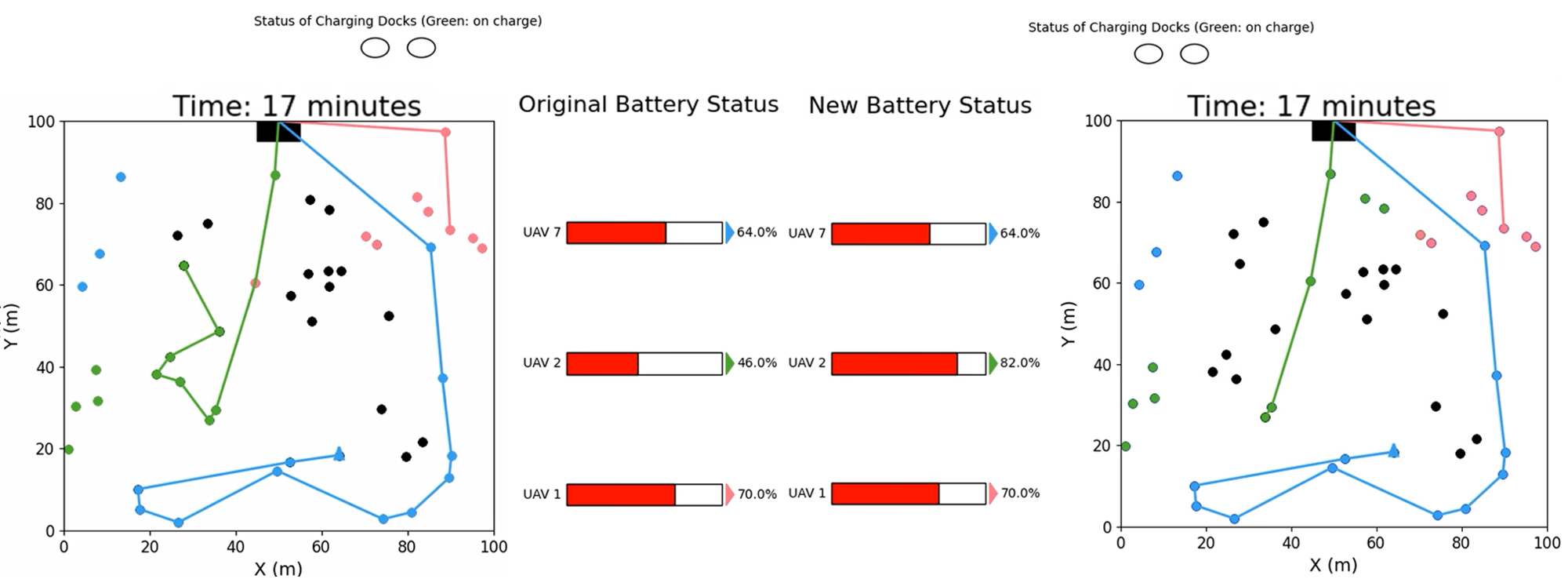}
        \caption{Snapshot at 17 minutes: All robots are airborne, maximizing operational efficiency}
        \label{fig:subim3}
    \end{subfigure}
    \hspace{5pt}
    \begin{subfigure}{0.48\textwidth}
        \includegraphics[width=\linewidth]{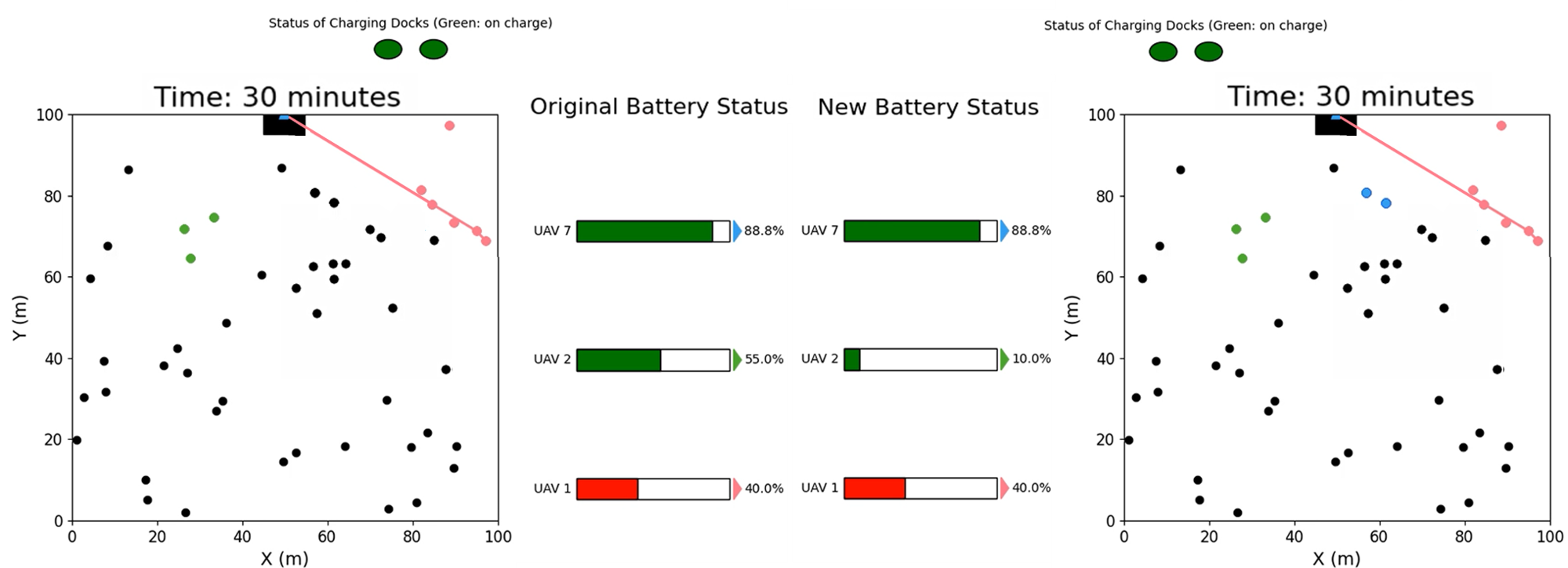}
        \caption{Snapshot at 30 minutes: Robots 1 continue their missions, while 2 and 7 recharging}
        \label{fig:subim4}
    \end{subfigure}
    \caption{Sequential snapshots from the simulation demonstrate the effectiveness of the scheduling algorithm at different stages of the robot operation cycle. The algorithm optimally selected and scheduled 3 out of 7 available robots (UAV1, UAV2, and UAV7), maximizing total flying time during the scheduling horizon with only 2 charging stations available. The figure on the left shows the simulation based on the original schedule, while the figure on the right corresponds to the simulation with the new schedule, adjusted for a 2-minute delay of UAV2.}
    \label{fig:combined}
\end{figure*}

\subsection{Finding a new scheduling horizon with safety margins}\label{Scheduling_Horizon}

In this section, we demonstrate the effectiveness of our proposed method for to reduce the overall scheduling horizon as a function of the safety margin parameter $\epsilon$. We conducted experiments using a sample of 10 robots with random cycle times. The parameter $\epsilon$ is set at $10\%$ for each robot. The table below presents the results of 15 such instances.
\begin{table}[h]
\centering

\caption{Scheduling Horizon with 10\% $\epsilon$ value}
\label{tab:experiment-results}
\begin{tabular}{ccc} 

\toprule 
\textbf{Instance} & \textbf{Scheduling Horizon} & \textbf{Modified Scheduling Horizon} \\ 
\midrule
        1 & 11592 & 504 \\ 
        2 & 4620 & 504 \\
        3 & 1287 & 540 \\ 
        4 & 5148 & 1287 \\
        5 & 9900 & 900 \\ 
        6 & 11310 & 504 \\ 
        7 & 5544 & 504 \\
        8 & 5400 & 540 \\ 
        9 & 7920 & 792 \\
        10 & 2700 & 540 \\ 
        11 & 5544 & 3360 \\ 
        12 & 3192 & 2128 \\ 
        13 & 4200 & 2970 \\
        14 & 1440 & 1296 \\
        15 & 2160 & 1188 \\

\bottomrule 

\end{tabular}\label{Table1}

\end{table}

As evidenced by the numbers above, scheduling robots with their maximum operational time can lead to a long scheduling horizon, and present computational challenges. However, incorporating a safety margin of up to 10\% in the flying time can notably decrease the overall scheduling horizon and computational time for finding a schedule. Fig.\ref{Computation Time} illustrates the effectiveness of reducing computation time.

\section{Simulation results}\label{Sim}

We now present the results of simulations conducted to validate the effectiveness of  a) selecting and scheduling robots when charging resources are limited, b) reducing the scheduling horizon along with incorporating a safety margin in flight time, and c) demonstrating robustness when a robot misses its scheduled charging time. These simulations involve selecting among 7 robots with 2 charging stations. Detailed parameters regarding simulations can be found in the   \href{https://github.com/Nitesh-mk/Persistent-Scheduling-Problem-Data-Set.git}{GitHub Repository}.

 The simulations were executed on a computational platform comprising an Ubuntu 22.04 NUC computer, equipped with an Intel i7-6770HQ processor and 32GB of RAM. The simulation environment replicated a realistic operational scenario where multiple robots were required to complete a series of tasks while managing their energy consumption effectively.

\begin{itemize}
    \item All robots depart from a designated charging spot at coordinates (50,100).
     \item The total number of available robots is 7, parameters details can be found from the given link: \href{https://github.com/Nitesh-mk/Persistent-Scheduling-Problem-Data-Set.git}{GitHub Repository}.
    \item Robots were tasked with visiting predetermined locations, represented by random 30 vertices, to simulate a typical surveillance or delivery mission.
    \item The robots operate at an average speed of 16 m/s, powered by 4000mAh 22.2V Li-po batteries.  
    \item The duration of time slot chosen for scheduling is $1$ minute.
    \item The path planning relied on the meta-heuristic team orienteering problem, which was implemented to determine the feasible sequence of visits. \cite{lee2024meta}

\end{itemize}

The algorithm selected 3 robots (UAV1, UAV2, UAV7) from the 7 available robots, with safety margins of 1, 2, and 4 minutes respectively. This selection provided a total flying time of 58 minutes within a cycle time of 36 minutes and effectively scheduled the robots on 2 charging stations. The schedule of the robots is shown in Fig.\ref{fig:charging_schedule}. The green shaded region represents the time when the robot needs to secure the charging station, and the red region is the time when the robot is available for the mission. This schedule repeats every 36 minutes.

We also considered the case when a robot misses its deadline to secure the charging station. As seen in Fig.\ref{fig:charging_schedule}, the robot (UAV2) needs to secure the charging station at 1 minute. However, if, due to unforeseen circumstances, the robot reaches the charging station at 3 minutes, the situation changes. UAV2 has a safety margin of 2 minutes, so reaching late is not immediately critical. However, providing a charging station at that time might disrupt the schedule of other robots. Using the discussion from Section \ref{rob}, we found that the earliest available charging station at 3 minutes can be given according to the new schedule in Fig.\ref{fig:charging_schedule}, compared to 10 minutes as per the original schedule without disrupting the cycle of other robots.

To visualize the newly computed schedule,  Fig.\ref{fig:combined} displays snapshots from the simulation at key moments: the 4th, 11th, 17th, and 30th slots. The circular icons at the top of each image indicate the status of the charging stations. Initially, at the 4-minute time slot, robot 2 is shown on the charging pads, strategically synchronized with robots 1 and 7, who are in the middle of their missions. As the simulation progresses, it is evident that no more than 2 charging stations are required at any given time to continue the mission. The feasible sequence of visits by the robots is determined using the meta-heuristic team orienteering problem discussed in \cite{lee2024meta}, showcasing the algorithm's adaptability to existing literature.


\section{CONCLUSIONS}\label{Con}

This paper presents a framework for efficiently utilizing charging stations by staggering the robots for recharging  to achieve long-term autonomy. It discusses the practical implementation of the algorithm by introducing the concept of a reduced scheduling horizon, which also provides a safety margin for the robots. The paper includes simulation results for a persistent surveillance problem to demonstrate the ease of extending this algorithm to different existing works in the literature. Additionally, the paper addresses robustness analysis, highlighting how the system adapts when a robot fails to secure a charging station on time. Overall, this paper covers the entire framework from scheduling to safety margins to robustness analysis, showing its adaptability and potential for integration into existing literature.

Currently, this method primarily focuses on cases where robots are fully charged and discharged. A natural extension of this work would be minimizing the resources required when partial charging and discharging is allowed.

The general problem involves combining routing of battery-limited robots with scheduling their recharging at the charging stations. The adopted approach decouples the routing problem from the scheduling problem in the following way: the operational time becomes a constraint for routing a robot so that all points of interest are covered within the smallest number of cycles. 






\bibliographystyle{IEEEtran}
\bibliography{ref}

\end{document}